\documentclass[twocolumn,a4paper]{article}
\usepackage[utf8]{inputenc}
\usepackage[T1]{fontenc}
\usepackage{textcomp}

\usepackage[a4paper,margin=2cm]{geometry}
\usepackage{booktabs}
\usepackage{caption}

\usepackage[backend=biber,style=vancouver]{biblatex}
\usepackage{csquotes}

\usepackage{xcolor}
\usepackage{listings}
\usepackage[most]{tcolorbox}
\usepackage{pdflscape}

\usepackage{hyperref}
\hypersetup{colorlinks=false,pdfborder={0 0 0}}

\lstdefinestyle{rawtextstyle}{
    backgroundcolor=\color{black!5},
    basicstyle=\ttfamily\footnotesize,
    breaklines=true,
    postbreak=\mbox{\textcolor{red}{$\hookrightarrow$}\space},
    keepspaces=true
}

\title{Telenor Nordics Customer Service Self-Help Corpus}

\author{Mike Riess\\[3pt]
\normalsize Research and Innovation, Telenor Group, Oslo, Norway\\
\normalsize E-mail any correspondence to: mike.riess@telenor.com}
\date{}

\begin{document}

\twocolumn[
\begin{@twocolumnfalse}
\maketitle
\begin{abstract}
\noindent This paper presents a multilingual customer service self-help corpus comprising 1,122 manually validated documents in Finnish, Danish, Norwegian, and Swedish, totaling 274,599 words and 1,884,833 characters. The documents have been sourced from the public self-help pages of four Nordic telecommunications operators and subsequently filtered for person-identifiable information and relevance through a combined LLM and human annotation pipeline. Domain-specific datasets for Nordic languages remain scarce, particularly in customer service: a domain of growing importance for retrieval-augmented generation, cross-lingual transfer learning, and emerging agent-based service architectures. An analysis of the corpus reveals substantial variation in document length and structure across operators, reflecting distinct editorial strategies, as well as broad topical coverage spanning network hardware, mobile services, TV and streaming, billing, and account management. The dataset is publicly available under a CC-BY-NC-SA-4.0 license at \url{https://zenodo.org/records/20732652}, intended to support reproducible research in Nordic NLP and information retrieval.

\end{abstract}
\bigskip
\end{@twocolumnfalse}
]

\section{Introduction and context}
Telenor Group is a telecommunications provider with presence in Asia and the following Nordic countries: Norway, Finland, Sweden and Denmark. Aiming to offer the best possible customer service, research in this domain is of strategic importance to Telenor. Prior work has shown that customer satisfaction is closely associated with the quality of support provided \cite{Kvale2021}, underscoring the need for high-quality knowledge resources that can support both human agents and automated systems. Recent work on zero-shot classification of Norwegian customer service dialogues \cite{riess-jorgensen-2025-brage} further highlights the potential of large language models in this domain, as well as the need for domain-specific evaluation resources.

To help researchers base their work on real world data in an ethical and legal manner, we are sharing a corpus of 1,122 customer service self-help documents in Norwegian, Finnish, Swedish and Danish. The data has been sourced from the respective operators' public websites, and subsequently been manually annotated and filtered, to ensure that no Person-Identifiable Information (PII) is present in the data. The data collection and processing has been financed by Telenor Group, with the aim of aiding academic research and boosting AI innovation. In accordance with the guidelines in \cite{Gebru2021}, this report will document the complete process of creating the self-help corpus.


\section{Data collection and preparation}

The data preparation pipeline consists of several steps that transform raw web data into a structured dataset where irrelevant content or PII is removed. Below is an overview which will be discussed in detail in the following sections.

\begin{itemize}
    \item \textbf{Step 1: Web Scraping}
    The process begins by collecting raw HTML data from selected web pages using a specialized scraper.

    \item \textbf{Step 2: Data Preparation}
    The collected HTML files are converted to Markdown format. Unnecessary elements such as navigation menus, headers, and footers are removed using site-specific XPath rules to isolate the primary content.

    \item \textbf{Step 3: LLM Annotation}
    An LLM (Gemma-3-27b-it \cite{GemmaTeam2025}) is used to annotate the Markdown documents with further information (content relevance, PII detection, text span), which will be validated by the human annotator later.

    \item \textbf{Step 4: LLM Translation}
    The Markdown documents are machine-translated into English to reduce translation errors by the human annotator. The text was translated using Gemma-3-27b-it, and the prompt used for translation can be seen in appendix 2.

    \item \textbf{Step 5: Human Validation and Annotation}
    The automatically generated annotations are reviewed and corrected by a human annotator via a web interface (see appendix 3).
    
    \item \textbf{Step 6: Data filtering}
    The final step in the process is to filter out the documents that did not adhere to the relevance or legal criteria.
\end{itemize}

\begin{table*}[t]
\centering
\caption{Number of documents remaining after filtering steps for each website.} 
\begin{tabular}{lrrrrr}
\toprule
\textbf{Filtering Step} & \textbf{DNA FI} & \textbf{Telenor DK} & \textbf{Telenor NO} & \textbf{Telenor SE} & \textbf{Total} \\
\midrule
Initial documents        & 382 & 234 & 187 & 449 & 1{,}252 \\
Excluded: Contains PII   &   1 &   0 &   0 &   0 &    1 \\
Excluded: Not customer service & 12 & 25 & 1 & 17 & 55 \\
Excluded: Not self-help  &   2 &   0 &   2 &  24 &   28 \\
Excluded: Duplicates     &   4 &  30 &   8 &   3 &   45 \\
Excluded: Empty content  &   1 &   0 &   0 &   0 &    1 \\
\midrule
Total excluded           &  20 &  55 &  11 &  44 &  130 \\
\textbf{Total accepted}  & \textbf{362} & \textbf{179} & \textbf{176} & \textbf{405} & \textbf{1{,}122} \\
\bottomrule
\end{tabular}
\label{tab:filtering_summary}
\end{table*}

\begin{table*}[h]
    \centering
    \caption{Description of dataset fields}
    \begin{tabular}{lll}
        \toprule
        \textbf{Field Name} & \textbf{Description} & \textbf{Example} \\
        \midrule
        source\_file\_relative\_path & Location of the source file (domain.xx/*) & tuki/-alykellot\_apple\-watch.html \\
        source\_domain & The website domain of the source file & www.dna.fi \\
        text & Markdown text from the document & ...\\
        language & The language of the source document & Finnish\\
        \bottomrule
    \end{tabular}
    \label{tab:fields}
\end{table*}

\subsection{Source data retrieval, annotation and filtering}\label{sec:data_retrieval}

The data source is the self-help pages of four telecommunications providers: Telenor Denmark\footnote{\url{https://telenor.dk/kundeservice}}, Telenor Norway\footnote{\url{https://telenor.no/kundeservice}}, Telenor Sweden\footnote{\url{https://telenor.se/support}} and DNA in Finland.\footnote{\url{https://www.dna.fi/tuki}} The data was retrieved with permission from the operators and in accordance with their individual \textit{robots.txt} disallow patterns (none of which included the self-help pages). The date of scraping was 23/05/2025. Table~\ref{tab:filtering_summary} shows an overview of the number of retrieved HTML documents from each of the providers in its \textit{initial documents} row. Once retrieved, the HTML data is converted to Markdown and manually inspected and annotated. The initial documents have been annotated based on the following criteria:

\begin{enumerate}
    \item \textbf{Customer service:} The text is related to products and services with the operator.
    \item \textbf{No Person-Identifiable Information (PII):} The text does not contain any PII.
    \item \textbf{Self-help:} The text contains instructions or facts about products and services that can be used to solve problems and thus reduce the load on customer service.
    \item \textbf{Duplicates:} The text does not exist in the same form in more than one document.
\end{enumerate}

The annotation has been performed in two steps: 1) An initial annotation attempt was performed on the Markdown document with \textbf{Gemma-3-27b-it}, 2) A manual review and correction was subsequently performed by a human annotator. The prompt used for step one can be seen in appendix 1. After the annotation, the data was filtered based on the mentioned criteria. The number of documents before and after filtering can be found in the top and bottom rows of Table~\ref{tab:filtering_summary}. Across all the websites, a total of 130 documents were removed based on the filtering criteria. The total count of accepted documents is 1,122. To report corpus size in consistent linguistic units, each document has been tokenized into words using spaCy~\cite{Honnibal2020spacy} and the language-specific pipelines listed in Table~\ref{tab:tokenizers}. Word-counts are performed after removal of Markdown formatting to avoid any Markdown-specific influence on the counts. In addition, the raw character counts (after Markdown removal) are reported. Across all websites this yields 274{,}599 words and 1{,}884{,}833 characters.

\begin{table*}[t]
\centering
\caption{Tokenisation models used per language. spaCy~3.8 word-token pipelines, built on the Universal Dependencies treebanks~\cite{nivre-etal-2016-universal}, are used to count words per language.}
\label{tab:tokenizers}
\begin{tabular}{lll}
\toprule
\textbf{Language} & \textbf{spaCy pipeline (v3.8)} & \textbf{UD training treebank} \\
\midrule
Finnish (DNA FI)       & \texttt{fi\_core\_news\_sm} & UD Finnish-TDT~\cite{haverinen2014turku} \\
Danish (Telenor DK)    & \texttt{da\_core\_news\_sm} & UD Danish-DDT~\cite{buchkromann2003danish} \\
Swedish (Telenor SE)   & \texttt{sv\_core\_news\_sm} & UD Swedish-Talbanken~\cite{nivre2007talbanken} \\
Norwegian (Telenor NO) & \texttt{nb\_core\_news\_sm} & UD Norwegian-Bokmaal~\cite{solberg2014norwegian} \\
\bottomrule
\end{tabular}
\end{table*}

\subsection{Annotation quality}\label{sec:annotation_quality}

To assess the reliability of the LLM pre-annotation, the Gemma-3-27b-it \cite{GemmaTeam2025} annotations were compared with the human-reviewed versions across all 1,251 matched document pairs (Table~\ref{tab:agreement}). Agreement on the boolean classification fields was high: 95.5\% for customer service relevance, 93.5\% for self-help classification, and 99.8\% for PII detection. In all observed disagreements, the LLM predicted \textit{True} where the human annotator corrected to \textit{False}, indicating a positive bias. The LLM tended to be overly inclusive, classifying general information or navigation pages as customer service content. Typical examples include index pages, language selectors and contact pages.

The LLM successfully extracted content span boundaries in 74.9\% of documents, but exact span matches with the human annotator were rare (6.8\%). The LLM identified approximate content regions, which the human then refined based on editorial judgment about what constitutes actionable self-help content. The high classification agreement combined with low span agreement confirms that the LLM served its intended role as a pre-annotation accelerator rather than a replacement for human judgment.

\begin{table}[t]
\centering
\caption{Agreement between LLM (Gemma-3-27b-it) pre-annotation and human review across 1,251 matched document pairs.}
\label{tab:agreement}
\small
\setlength{\tabcolsep}{3pt}
\begin{tabular}{lrrr}
\toprule
\textbf{Field} & \textbf{Agreed} & \textbf{Total} & \textbf{Agreement} \\
\midrule
Customer service related & 1{,}195 & 1{,}251 & 95.5\% \\
Self-help resource       & 1{,}170 & 1{,}251 & 93.5\% \\
Contains PII             & 1{,}248 & 1{,}251 & 99.8\% \\
\midrule
Span extraction success  &     937 & 1{,}251 & 74.9\% \\
Exact span match         &      64 &     937 &  6.8\% \\
\bottomrule
\end{tabular}
\end{table}


\section{Dataset composition}

This section provides a descriptive analysis of the final dataset.

\subsection{Structure and fields}

The data is organized into separate folders for each of the source websites. Each folder consists of JSON-files, which in turn have the data fields shown in Table~\ref{tab:fields}.

\subsection{Statistics and distributions}

The dataset contains 274{,}599 words and 1{,}884{,}833 characters across all four languages. Table~\ref{tab:token_stats} presents a detailed breakdown of document lengths per language.

\begin{table*}[t]
\centering
\caption{Corpus size and document length by language, in linguistic units (spaCy word tokens) and characters. All counts are computed on Markdown-stripped text. Per-language tokenisation models are listed in Table~\ref{tab:tokenizers}.}
\label{tab:token_stats}
\begin{tabular}{lrrrrr}
\toprule
\textbf{Language} & \textbf{Docs} & \textbf{Words} & \textbf{Avg Words/Doc} & \textbf{Median Words/Doc} & \textbf{Characters} \\
\midrule
DNA FI      & 362 & 87{,}306 & 241 & 164 & 749{,}435 \\
Telenor DK  & 179 & 44{,}744 & 250 & 196 & 268{,}032 \\
Telenor NO  & 176 & 72{,}749 & 413 & 320 & 445{,}850 \\
Telenor SE  & 405 & 69{,}800 & 172 &  99 & 421{,}516 \\
\midrule
\textbf{Total} & \textbf{1{,}122} & \textbf{274{,}599} & \textbf{245} & \textbf{151} & \textbf{1{,}884{,}833} \\
\bottomrule
\end{tabular}
\end{table*}

The dataset exhibits substantial variation in document length across the four operators. Norwegian documents are the longest by a clear margin (median 320 words, 1{,}958 characters) and Swedish the shortest (median 99 words, 591 characters), with Finnish and Danish in between. Finnish ranks mid-length by word count (median 164) but second-longest by characters (median 1{,}324), as its words are longer on average (8.6 characters per word, versus roughly 6 for the others). Both word and character counts are therefore reported, the latter giving a more comparable measure of length across languages.


The length variation reflects editorial differences between the operators. DNA Finland maintains extensive, detailed device-specific guides, including legacy hardware for backward compatibility, giving its documents the highest character counts after Norwegian. Telenor Sweden, by contrast, favors shorter, more focused articles: its five shortest documents contain only 12 to 18 words each, typically addressing a single troubleshooting step. Norwegian documents have the highest structural density, averaging 6.1 headings and 13.1 list items per document, suggesting a step-by-step instructional style. These differences are relevant for downstream tasks: retrieval systems must accommodate documents ranging from 10 to 1{,}789 words (77 to 10{,}437 characters), and chunking strategies may need to be adapted per operator.

\begin{figure}[h]
    \centering
    \includegraphics[width=0.9\linewidth]{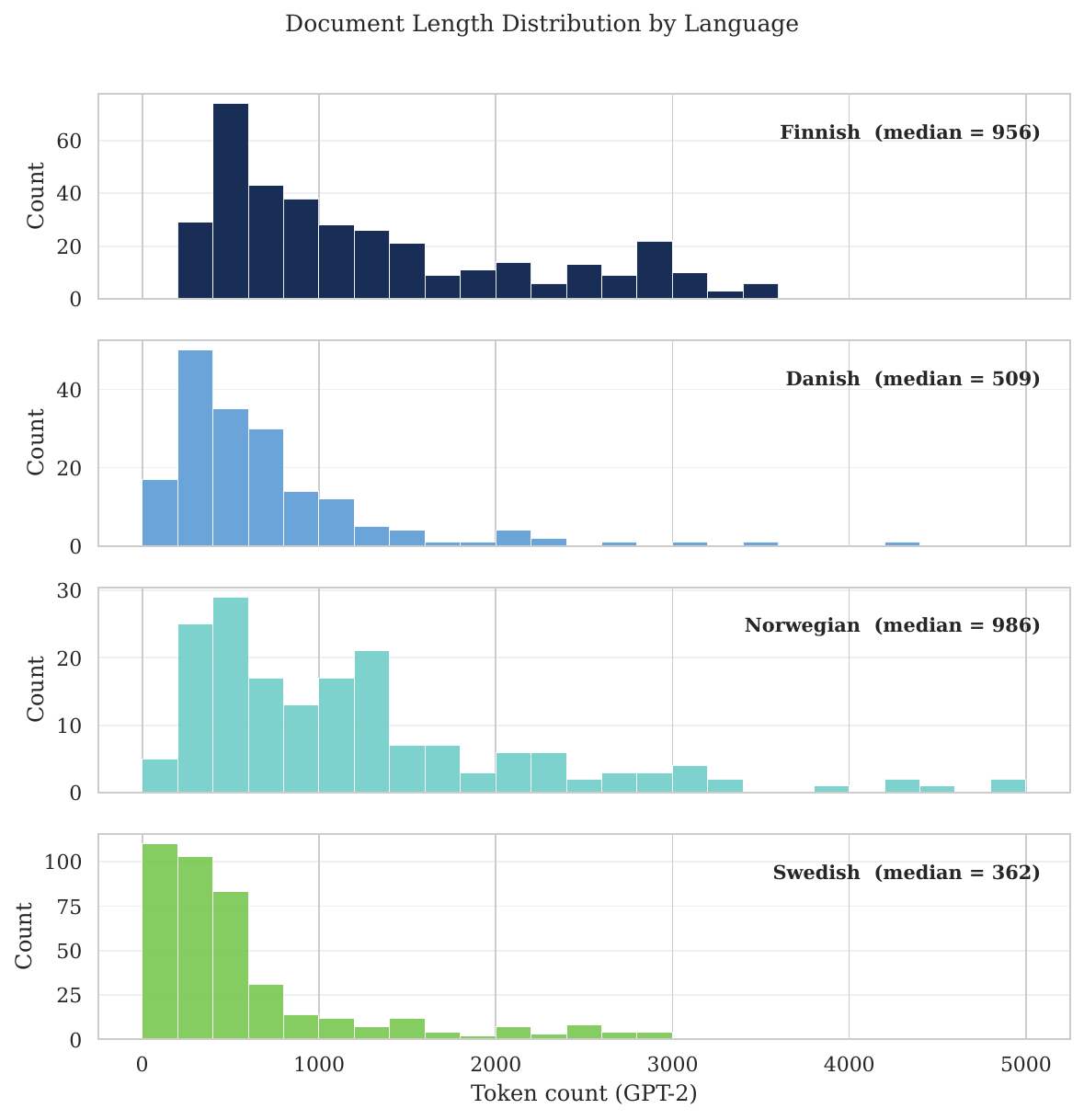}
    \caption{Distribution of document lengths (in words) across the corpus.}
    \label{fig:length_dist}
\end{figure}

\subsubsection{Topic analysis}

To characterize the topical composition of the corpus, zero-shot content classification using \textbf{multilingual-e5-large} \cite{Wang2024} was performed. Each document is assigned the category, among ten predefined categories, whose embedding has the highest cosine similarity to the document. The category prompts are applied in the document's individual language (an English prompt template can be seen in appendix 5), together with the original language version of the text, to avoid any translation-related bias. The prompts were translated into Finnish, Danish, Norwegian and Swedish and verified by a native speaker of each language (appendix 6), so that prompt and document share the same language.

The corpus covers a broad range of telecommunications customer service topics. Router and network hardware documentation constitutes the largest category (377 documents, 34\%), the majority of them Finnish (218), reflecting DNA Finland's practice of maintaining detailed setup guides for both current and legacy equipment: Finnish likewise dominates network and coverage content (74 of 114 documents). TV and streaming (163 documents) and mobile subscriptions and services (128) are the next largest categories, both led by Swedish documents (104 and 73 respectively), consistent with Telenor Sweden's broader consumer portfolio.

The category distribution reflects genuine differences in the product catalogs and editorial strategies, rather than a uniform topical structure across languages. This is a relevant property for researchers: the dataset can support both within-category cross-lingual comparisons (e.g., how different operators document router setup) and cross-category analyses within a single language.

\begin{figure}[h]
    \centering
    \includegraphics[width=0.9\linewidth]{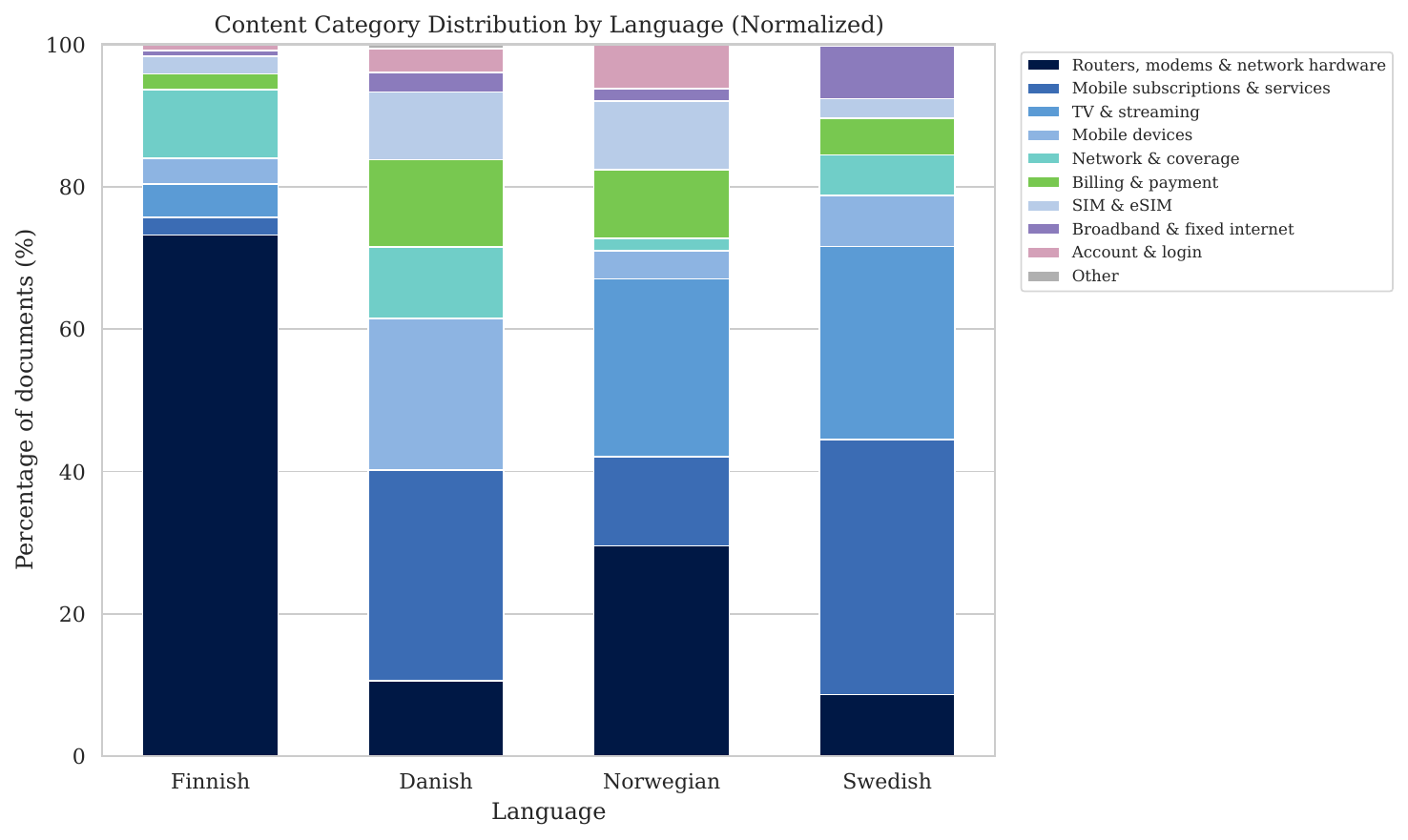}
    \caption{Normalized distribution of content categories across the corpus, based on zero-shot embedding classification of the original-language documents using native-language category prompts.}
    \label{fig:categories}
\end{figure}


\section{Discussion and usage}

This dataset is intended for research in natural language processing and information retrieval for Nordic languages, with a focus on the customer service domain. Below, the primary use cases, the unique properties of the corpus, and planned future work are outlined.

\subsection{Intended use}

The corpus can serve as a foundation for several types of NLP tasks. As a knowledge base for retrieval-augmented generation (RAG), the documents provide realistic, domain-specific content for building and evaluating Nordic-language QA systems. The multilingual coverage within a single domain makes the dataset suitable for cross-lingual transfer learning experiments, where models trained on one language can be evaluated on another. The dataset can also be used for evaluating embedding models on Nordic languages, complementing existing benchmarks such as the Scandinavian Embedding Benchmark \cite{Enevoldsen2024}. Finally, the natural problem-solution structure of self-help documents makes them well suited as test data for emerging agent-based service architectures, where systems must retrieve, interpret, and act on procedural information.

\subsection{What makes this dataset distinct}

Most existing customer service datasets are either English-language, synthetically generated, or based on chat logs and tickets. Recent benchmarks such as WixQA \cite{Cohen2025}, which provides enterprise self-help articles with QA pairs from Wix.com, and CXMArena \cite{CXMArena2025}, which targets customer experience management tasks including knowledge base refinement and multi-turn RAG, have advanced evaluation of customer service NLP systems. However, both are English-language and not specific to telecommunications. To the best of the author's knowledge, no comparable corpus exists for Nordic languages in this domain.

This corpus differs from existing resources in several respects: the documents are real production content from active telecommunications operators, manually validated for quality and PII, multilingual within the same domain, and freely available under an open license. The combination of these properties makes it possible to evaluate NLP systems under realistic conditions for Nordic languages, where domain-specific datasets remain scarce.

The four operators, while part of the same group, operate as separate companies with distinct product catalogs, editorial strategies, and customer bases. This introduces natural variation in document length, structure, and topical emphasis. These are properties that are representative of real-world multilingual deployment scenarios but rarely captured in curated benchmarks.

\subsection{Future Work}

A planned extension is the development of a companion evaluation set with queries and ground-truth document relevance judgments, enabling benchmarking of retrieval systems on this corpus.


\subsection{Limitations}

The dataset has several limitations that users should be aware of:

The corpus is a static snapshot from May 2025. Customer service content is updated regularly as products and services change, and parts of the dataset may already be outdated at the time of publication.

All annotations were performed by a single annotator. While the LLM pre-annotation agreement rates provide some measure of annotation consistency, inter-annotator agreement cannot be assessed. Furthermore, the human annotator reviewed the LLM predictions before making corrections, which may introduce anchoring bias \cite{Tversky1974}: a tendency to accept the LLM's judgment unless it is obviously incorrect. Recent work has shown that annotators who review LLM suggestions tend to adopt them at high rates, significantly shifting label distributions compared to independent annotation \cite{schroeder-etal-2025-loop}.

The annotation pipeline relies on machine translation: documents were translated into English with Gemma-3-27b-it to assist the annotator. Translation quality was not formally evaluated, and this may have influenced the quality of the annotation. The annotator is fluent in Danish and Norwegian but fully relied on the English translations for the Finnish documents, and to a lesser degree on the Swedish ones. Translation errors may therefore have influenced those annotations more than the Danish and Norwegian ones. Users of this corpus should be aware of this.

The dataset does not include question-answer pairs or evaluation queries. Developing such a companion resource is planned future work.

The language distribution is unbalanced, with fewer Danish (179) and Norwegian (176) documents than Finnish (362) and Swedish (405). Depending on the task, the smaller subcorpora may not provide sufficient data for robust per-language evaluation.

Finally, the CC-BY-NC-SA-4.0 license restricts commercial use of the dataset. This is a deliberate choice to protect the operators' content while enabling academic research.


\subsection{Availability and License}

The dataset is archived at Zenodo \cite{Riess2026dataset} under a CC-BY-NC-SA-4.0 license. The repository at \url{https://github.com/tnresearch/tn_selfhelp_corpus} contains the filtering and analysis code, as well as a Docker configuration for reproducing all analyses presented in this paper.

\section*{Acknowledgements}

The author would like to thank three colleagues Mikko Ollgren, Eirik Fagerhaug, and Anders Bresell (native speakers of Finnish, Norwegian and Swedish, respectively) for verifying and correcting the category-prompt translations used in the topic analysis (Table A3).


\newpage
\printbibliography
\newpage


\appendix 
\onecolumn
\section{Appendix}
\renewcommand{\thefigure}{A\arabic{figure}}
\renewcommand{\thetable}{A\arabic{table}}
\setcounter{figure}{0}
\setcounter{table}{0}

\subsection{1. Prompt used for annotation}\label{sec:appendix_annotation_prompt}

\begin{tcolorbox}[
    breakable,
    colback=gray!10,
    colframe=black,
    fonttitle=\bfseries,
    title=Data annotation prompt: gemma-3-27b-it
]
\begin{lstlisting}[style=rawtextstyle]

You are an expert text analyst tasked with annotating web page content based on specific criteria. Analyze the following document text and provide your analysis ONLY in the form of a valid JSON object matching the specified structure.

**JSON Structure Required:**

```json
{
  "filtering": {
    "customer_service_related": boolean, // True if the text relates to customer service, support, or complaints.
    "self_help_resource": boolean // True if the text provides instructions, guides, or factual info about products/services.
  },
  "pii_detection": {
    "contains_pii": boolean // True if the text contains any full names of individuals (e.g., "John Doe").
  },
  "content_selection": {
    "selected_span": {
      "starting_substring": string | null, // The first ~5-10 words of the selected text block (including the heading if there is one). null if no relevant block found.
      "ending_substring": string | null, // The last ~5-10 words of the selected text block. null if no relevant block found.
      "heading": string | null // A brief, concise summary in English of the page title or main H1 heading. null if none found.
    }
  }
}
```

**Analysis Task:**

1.  **Filtering:** Determine the boolean values for `customer_service_related` and `self_help_resource`.
2.  **PII Detection:** Determine if the document contains any **full names of individuals**. Set `contains_pii` to true ONLY if full names are found.
3.  **Content Selection:** Identify the single, continuous block of text within the document that represents the most relevant core content for the purpose of customer service or self-help information. Exclude common headers, footers, navigation, ads, and boilerplate.
    *   If a relevant block is found:
        *   Identify the **first 5-10 words** of the block and provide them exactly as `starting_substring`.
        *   Identify the **last 5-10 words** of the block and provide them exactly as `ending_substring`.
        *   Generate a concise summary in English of the page's primary title or main heading and provide it as `heading`.
    *   If NO relevant block can be identified, set `starting_substring`, `ending_substring`, and `heading` all to `null`.

**Input Document Text:**
```text
{{DOCUMENT_TEXT}}
```

**Output (JSON only):**
```json
{
  "filtering": {
    "customer_service_related": ...,
    "self_help_resource": ...
  },
  "pii_detection": {
    "contains_pii": ... // Focused on full names
  },
  "content_selection": {
    "selected_span": {
      "starting_substring": "The first few words...", // Ensure this is **exactly** verbatim
      "ending_substring": "...the last few words.", // Ensure this is **exactly** verbatim
      "heading": "Concise English Page Title Summary..." // English summary of title/H1
    }
  }
}
```

\end{lstlisting}
\end{tcolorbox}

\subsection{2. Prompt used for translation}\label{sec:appendix_translation_prompt}

\begin{tcolorbox}[
    breakable,
    colback=gray!10,
    colframe=black,
    fonttitle=\bfseries,
    title=Data translation prompt: gemma-3-27b-it
]
\begin{lstlisting}[style=rawtextstyle]
Translate the following Markdown document into English. Preserve the original Markdown formatting (headings, lists, code blocks, links, bold, italics, etc.) exactly. Output ONLY the translated English Markdown text.

DOCUMENT_TEXT:
{{DOCUMENT_TEXT}}

TRANSLATED_ENGLISH_Markdown: 
\end{lstlisting}
\end{tcolorbox}

\newpage
\subsection{3. Annotation UI}\label{sec:appendix_annotation_ui}

\begin{figure}[h!]
    \centering
    \includegraphics[width=1\linewidth]{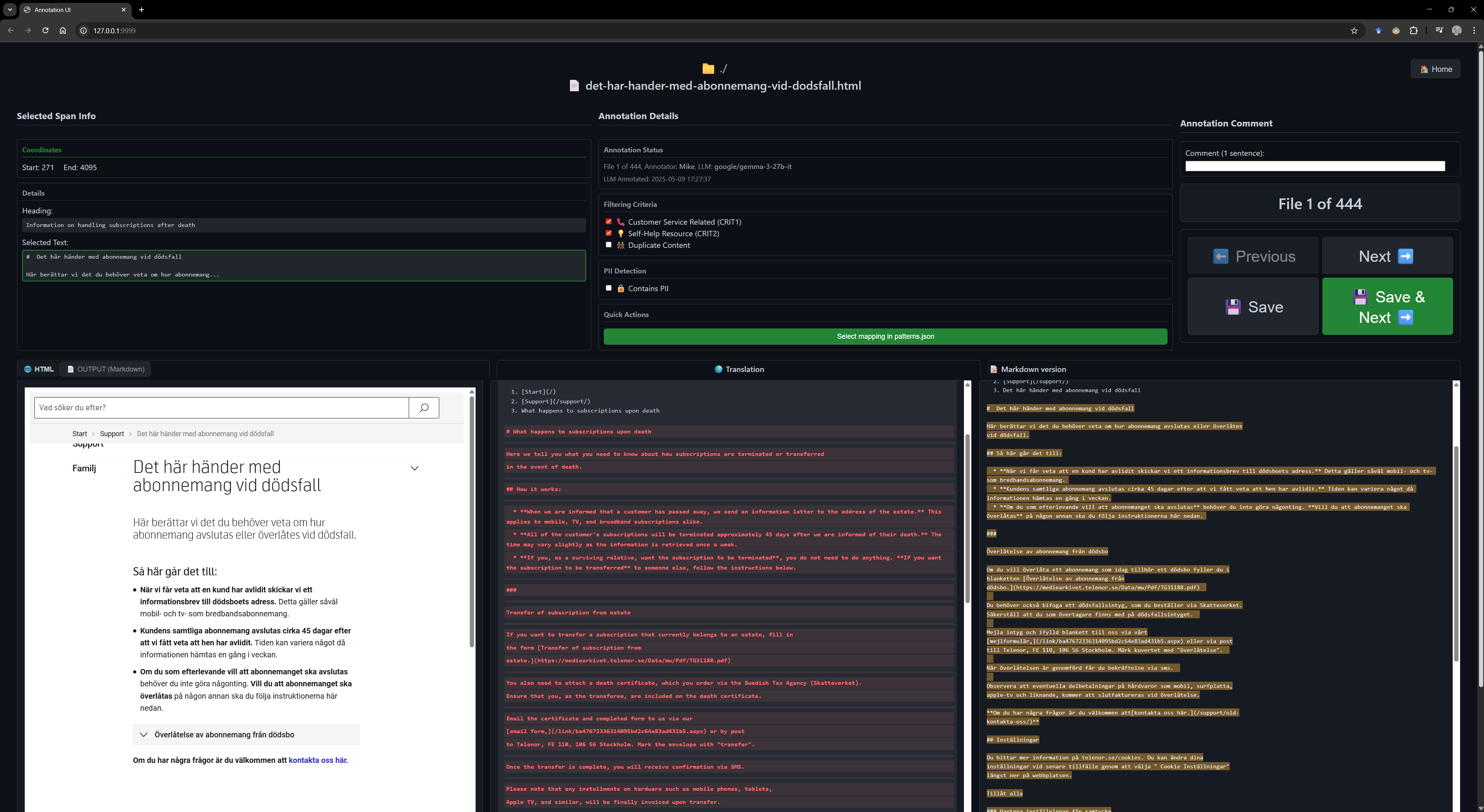}
    \caption{Screenshot of the Annotation UI used to select the relevant text and annotate the documents.}
    \label{fig:annotation_ui}
\end{figure}

Annotation and human review were carried out in a purpose-built web interface (Figure~\ref{fig:annotation_ui}). The tool is an internal solution and is not released together with the dataset, but will be described here so that the annotation procedure can be understood and, in part, reproduced with other tooling. For each document the interface presents three synchronised views: the full original-language HTML page as served to end users, its machine-translated English version (center), and the original document text as Markdown (right). The LLM pre-annotations, the filtering flags, PII detection and the suggested content span are displayed alongside these views, and the annotator can adjust the pre-marked text span (accepting, extending, or replacing the LLM suggestion) in order to quality-assure the selected content. The UI included the ability to use site-specific regex-patterns for span selection in the cases where the LLM-suggested span would be invalid.

\newpage
\subsection{4. Content Categories}\label{sec:appendix_categories}

\begin{table}[h]
\centering
\caption{Content category distribution by language, based on zero-shot embedding classification of the documents using native-language category prompts.}
\label{tab:categories}
\begin{tabular}{lrrrrr}
\toprule
\textbf{Category} & \textbf{FI} & \textbf{DK} & \textbf{NO} & \textbf{SE} & \textbf{Total} \\
\midrule
Routers, modems \& network hardware & 218 & 34 & 55 & 70 & 377 \\
TV \& streaming & 16 & 0 & 43 & 104 & 163 \\
Mobile subscriptions \& services & 11 & 26 & 18 & 73 & 128 \\
Network \& coverage & 74 & 16 & 3 & 21 & 114 \\
Account \& login & 5 & 42 & 27 & 26 & 100 \\
Mobile devices & 13 & 14 & 2 & 38 & 67 \\
SIM \& eSIM & 13 & 15 & 10 & 25 & 63 \\
Broadband \& fixed internet & 1 & 15 & 8 & 28 & 52 \\
Billing \& payment & 5 & 10 & 7 & 13 & 35 \\
Other & 6 & 7 & 3 & 7 & 23 \\
\midrule
\textbf{Total} & \textbf{362} & \textbf{179} & \textbf{176} & \textbf{405} & \textbf{1{,}122} \\
\bottomrule
\end{tabular}
\end{table}

\subsection{5. Descriptions used for topic classification}
\label{sec:appendix_classification}

\begin{table}[h]
\centering
\caption{Category descriptions used for zero-shot
embedding classification.}
\label{tab:appendix_descriptions}
\begin{tabular}{p{4cm}p{10cm}}
\toprule
\textbf{Category} & \textbf{Description} \\
\midrule
Mobile subscriptions \& services & Mobile phone subscriptions, plans, prepaid, roaming, MMS, SMS, VoLTE, VoWiFi, mobile services \\
Broadband \& fixed internet & Broadband, fiber, DSL, fixed internet connection, internet speed, home internet \\
Routers, modems \& network hardware & Router setup, modem configuration, WiFi settings, network equipment, dongles, repeaters \\
TV \& streaming & Television services, streaming, TV boxes, channels, remote control, set-top box \\
Billing \& payment & Invoice, billing, payment, pricing, charges, fees, credit, subscription cost \\
Account \& login & My account, login, password, profile, app, self-service portal \\
SIM \& eSIM & SIM card, eSIM, activation, SIM swap, PIN, PUK \\
Mobile devices & Smartphones, tablets, smartwatches, phone setup, device troubleshooting, screen, insurance \\
Network \& coverage & Network coverage, outages, 4G, 5G, signal strength, connectivity problems \\
Other & General customer service information, terms and conditions, contact information, moving, address change, security \\
\bottomrule
\end{tabular}
\end{table}

\newpage
\subsection{6. Per-language category prompts}
\label{sec:appendix_prompts}

The primary topic classification (Section~3, Figure~\ref{fig:categories}) uses category prompts in each document's own language. Each prompt is the full English template of Table~\ref{tab:appendix_descriptions} (category name and key terms) translated into the language and verified by a single native speaker: the prompts are listed in Table~\ref{tab:category_prompts_multilingual}.

\begin{landscape}
\begin{table}[p]
\centering
\footnotesize
\setlength{\tabcolsep}{4pt}
\caption{Per-language category prompts used for the primary topic classification. Each cell is the full prompt (translated category name and key terms) embedded as the ``query'' for that language. The Finnish, Danish, Norwegian and Swedish prompts were translated and verified by a native speaker of each language (the author is a native Danish speaker and thus translated the Danish prompt). The English prompts are the Table~\ref{tab:appendix_descriptions} descriptions with the category name prepended.}
\label{tab:category_prompts_multilingual}
\begin{tabular}{p{3.0cm}p{4.6cm}p{4.6cm}p{4.6cm}p{4.6cm}}
\toprule
\textbf{Category} & \textbf{Finnish} & \textbf{Danish} & \textbf{Norwegian} & \textbf{Swedish} \\
\midrule
Mobile subscriptions \& services & Matkapuhelinliittymät ja -palvelut: matkapuhelinliittymät, liittymäpaketit, prepaid-liittymät, verkkovierailu, MMS, SMS, VoLTE, VoWiFi, matkapuhelinpalvelut & Mobilabonnementer og tjenester: mobilabonnementer, pakker, taletidskort, roaming, MMS, SMS, VoLTE, VoWiFi, mobiltjenester & Mobilabonnement og tjenester: mobilabonnement, pakker, kontantkort, roaming, MMS, SMS, VoLTE, VoWiFi, mobiltjenester & Mobilabonnemang och tjänster: mobilabonnemang, paket, kontantkort, roaming, MMS, SMS, VoLTE, VoWiFi, mobiltjänster \\
\addlinespace
Broadband \& fixed internet & Laajakaista ja kiinteä internet: laajakaista, valokuitu, DSL, kiinteä internetyhteys, internetnopeus, kotinetti & Bredbånd og fast internet: bredbånd, fiber, DSL, fast internetforbindelse, internethastighed, hjemmeinternet & Bredbånd og fast internett: bredbånd, fiber, DSL, fast internettforbindelse, internetthastighet, hjemmeinternett & Bredband och fast internet: bredband, fiber, DSL, fast internetanslutning, internethastighet, hemmainternet \\
\addlinespace
Routers, modems \& network hardware & Reitittimet, modeemit ja verkkolaitteet: reitittimen asennus, modeemin konfigurointi, WiFi-asetukset, verkkolaitteet, USB-adapterit, toistimet & Routere, modems og netværkshardware: router opsætning, modem konfiguration, WiFi indstillinger, netværksudstyr, dongle, repeater & Rutere, modemer og nettverksutstyr: ruter oppsett, modem konfigurasjon, WiFi innstillinger, nettverksutstyr, dongel, forsterker & Routrar, modem och nätverkshårdvara: routerinstallation, modemkonfiguration, WiFi-inställningar, nätverksutrustning, donglar, repeaters \\
\addlinespace
TV \& streaming & TV ja suoratoisto: televisiopalvelut, suoratoisto, TV-mediatoistin, kanavat, kaukosäädin, digiboksi & TV og streaming: tv-tjenester, streaming, tv-bokse, kanaler, fjernbetjening, set-top boks & TV og streaming: tv-tjenester, streaming, tv-bokser, kanaler, fjernkontroll, set-top boks & TV och streaming: tv-tjänster, streaming, tv-boxar, kanaler, fjärrkontroll, set-top box \\
\addlinespace
Billing \& payment & Laskutus ja maksut: lasku, laskutus, maksu, hinnoittelu, maksut, palkkiot, hyvitys, tilausmaksu & Fakturering og betaling: faktura, fakturering, betaling, priser, afgifter, gebyrer, kredit, abonnementspris & Fakturering og betaling: faktura, fakturering, betaling, priser, avgifter, gebyrer, kreditt, abonnementspris & Fakturering och betalning: faktura, fakturering, betalning, priser, avgifter, gebyr, kredit, abonnemangskostnad \\
\addlinespace
Account \& login & Tili ja kirjautuminen: oma tili, kirjautuminen, salasana, profiili, sovellus, itsepalveluportaali & Konto og login: min konto, login, adgangskode, profil, app, selvbetjeningsportal & Konto og log inn: min konto, logg inn, passord, profil, app, selvbetjeningsportal & Konto och login: mitt konto, login, lösenord, profil, app, självbetjäningsportal \\
\addlinespace
SIM \& eSIM & SIM-kortti ja eSIM: SIM-kortti, eSIM, aktivointi, SIM-kortin vaihtaminen, PIN-koodi, PUK-koodi & SIM og eSIM: SIM-kort, eSIM, aktivering, SIM-kort skift, PIN, PUK & SIM og eSIM: SIM-kort, eSIM, aktivering, bytte av SIM-kort, PIN, PUK & SIM och eSIM: SIM-kort, eSIM, aktivering, SIM-kortsbyte, PIN, PUK \\
\addlinespace
Mobile devices & Mobiililaitteet: älypuhelimet, tabletit, älykellot, puhelimen käyttöönotto, laitteiden vianmääritys, näyttö, vakuutus & Mobile enheder: smartphones, tablets, smartwatches, telefonopsætning, enhedsfejlfinding, skærm, forsikring & Mobile enheter: smarttelefon, nettbrett, smartklokke, telefonoppsett, enhetsfeilsøking, skjerm, forsikring & Mobila enheter: smartphones, tablets, smartwatches, telefoninställning, enhetsfelsökning, skärm, försäkring \\
\addlinespace
Network \& coverage & Verkko ja kattavuus: verkon kattavuus, häiriöt, 4G, 5G, signaalin voimakkuus, yhteysongelmat & Netværk og dækning: netværksdækning, driftsforstyrrelser, 4G, 5G, signalstyrke, forbindelsesproblemer & Nettverk og dekning: nettverksdekning, driftsforstyrrelser, 4G, 5G, signalstyrke, tilkoblingsproblemer & Nätverk och täckning: nätverkstäckning, driftstörningar, 4G, 5G, signalstyrka, anslutningsproblem \\
\addlinespace
Other & Muu: yleistä asiakaspalvelutietoa, käyttöehdot, yhteystiedot, muutto, osoitteenmuutos, tietoturva & Andet: generel kundeserviceinformation, vilkår og betingelser, kontaktinformation, flytning, adresseændring, sikkerhed & Annet: generell kundeserviceinformasjon, vilkår og betingelser, kontaktinformasjon, flytting, adresseendring, sikkerhet & Övrigt: allmän kundtjänstinformation, villkor, kontaktinformation, flytt, adressändring, säkerhet \\
\bottomrule
\end{tabular}
\end{table}
\end{landscape}

\end{document}